\documentclass[journal]{IEEEtran} 

\usepackage{amsmath,amssymb,amsfonts,bm,mathdots,amsthm}

\usepackage{graphicx}
\usepackage{epsfig}
\usepackage{epstopdf}
\usepackage{float}
\graphicspath{{Pictures/}{../jpeg/}} 

\usepackage{comment}

\usepackage{array}
\usepackage{multirow}
\usepackage{booktabs}
\usepackage{longtable}
\usepackage{adjustbox}
\usepackage{colortbl}

\usepackage{algorithm}
\usepackage{algpseudocode}

\ifCLASSOPTIONcompsoc
  \usepackage[caption=false,font=normalsize,labelfont=sf,textfont=sf]{subfig}
\else
  \usepackage[caption=false,font=footnotesize]{subfig}
\fi

\usepackage[dvipsnames]{xcolor}
\definecolor{lightblue}{RGB}{230,245,255}
\definecolor{skyblue}{RGB}{70,130,180}

\usepackage{cite}
\usepackage{hyperref}
\hypersetup{hidelinks=true}

\usepackage{gensymb}
\usepackage{textcomp}
\usepackage{pifont}
\usepackage{enumitem}
\usepackage{varioref}
\usepackage{latexsym}

\newcommand{\cmark}{\ding{51}}
\newcommand{\xmark}{\ding{55}}

\def\BibTeX{{\rm B\kern-.05em{\sc i\kern-.025em b}\kern-.08em
  T\kern-.1667em\lower.7ex\hbox{E}\kern-.125emX}}

\begin{document}
\title{A Fusion of context-aware based BanglaBERT and Two-Layer Stacked LSTM Framework for Multi-Label Cyberbullying Detection}


\author{%
 Mirza Raquib\IEEEauthorrefmark{1},
 Asif Pervez Polok\IEEEauthorrefmark{1},
 Kedar Nath Biswas, 
 Rahat Uddin Azad,
 Saydul Akbar Murad,
Nick Rahimi\IEEEauthorrefmark{2}

 \thanks{\IEEEauthorrefmark{1}These authors contributed equally.}
 \thanks{\IEEEauthorrefmark{2}Corresponding author.}%
 \thanks{Mirza Raquib: Department of Computer and Communication Engineering, International Islamic University Chittagong, Chattogram, Bangladesh; Department of Information and Communication Engineering, Noakhali Science and Technology University, Noakhali, Bangladesh. Email: mirzaraquib@iiuc.ac.bd}%
 \thanks{Asif Pervez Polok: mPower Social Enterprise. Email: asifpolok.research@gmail.com}%
 \thanks{Kedar Nath Biswas: Department of Information and Communication Engineering, Noakhali Science and Technology University, Noakhali, Bangladesh. Email: kedarnathbiswas06@gmail.com}%
 \thanks{Rahat Uddin Azad: Department of Software Engineering, Daffodil International University, Bangladesh. Email: rahatuddin786@gmail.com}%
 \thanks{Saydul Akbar Murad: School of Computing Sciences and Computer Engineering, University of Southern Mississippi, Hattiesburg, MS, USA. Email: saydulakbar.murad@usm.edu}%
 \thanks{Nick Rahimi: School of Computing Sciences and Computer Engineering, University of Southern Mississippi, Hattiesburg, MS, USA. Email: nick.rahimi@usm.edu}%
}

\maketitle
\begin{abstract}
Cyberbullying has become a serious and growing concern in today’s virtual world. When left unnoticed, it can have adverse consequences for social and mental health. Researchers have explored various types of cyberbullying, but most approaches use single-label classification. These methods assume that each comment contains only one type of abuse. In reality, a single comment may include overlapping forms such as threats, hate speech, and harassment. Therefore, multilabel detection is both realistic and essential. Multilabel cyberbullying detection has received limited attention, especially in low-resource languages like Bangla, where robust pre-trained models are scarce. Developing a generalized model with moderate accuracy remains challenging. Transformers offer strong contextual understanding but may miss sequential dependencies. Conversely, LSTM models capture temporal flow but lack semantic depth. We have proposed a robust fusion architecture that combines BanglaBERT-Large with a two-layer stacked LSTM. We have analyzed their behavior to jointly model context and sequence. The model is fine-tuned and evaluated on a publicly available dataset for multilabel cyberbullying: "Bangla multilabel cyberbully, sexual harassment, threat, and spam detection." We have applied different sampling strategies to tackle class imbalance. Evaluation uses multiple metrics, including accuracy, precision, recall, F1-score, Hamming loss, Cohen’s kappa, and AUC-ROC. We employ 5-fold cross-validation to assess the architecture's generalization. Experimental results show that our BanglaBERT-LSTM model achieves state-of-the-art performance with 94.31\% accuracy. It outperforms the prior state of the art by 0.67\% and consistently exceeds existing machine learning, deep learning, and hybrid techniques on this dataset. In addition, explainable AI techniques are integrated to improve model interpretability. Our work establishes a strong benchmark for multilabel cyberbullying detection and advances Bangla NLP in low-resource settings.

\end{abstract} 
\section{Introduction}
\label{sec:1}

The rapid growth of social media platforms has highlighted the need for robust NLP algorithms to effectively detect and analyze abusive language. Cyberbullying has emerged as a critical challenge on online platforms, significantly affecting users' mental well-being and social safety. Cyberbullying incidents are on the rise and can occur anywhere, at any time, due to the anonymity and lack of self-control offered by internet platforms \cite{OECD2024MentalHealth} \cite{GradesFixerCyberbullyingBangladesh}. Serious mental health issues, including anxiety, depression, and even suicidals, are common among victims \cite{Agustiningsih2024CyberbullyingMentalHealth}. Cyberbullying is reported by more than 56\% of teenagers \cite{Gazibara2023CyberbullyingMentalHealth}. Victims are more likely to experience worry, anxiety, and low self-esteem; some may even contemplate suicide \cite{Agustiningsih2024CyberbullyingMentalHealth}. In the field of cyberbullying detection, low-resource languages like Bangla are less explored than English and other high-resource languages. Bangla is one of the most widely spoken languages in the world, yet research on multilabel cyberbullying detection in Bangla remains very limited. Social media has become a major platform for communication, where user comments often include offensive language that can harm individuals and communities. Many earlier systems focused only on single-label detection. These models can only detect one abuse type at a time. But in reality, a single comment may include multiple types of abuse, such as harassment, hate speech, threats, or spam. For example, a comment may threaten a user and also include religious hate. A single-label model may detect only one and ignore the other. This can lead to missing dangerous content. In some cases, if a comment fails to get detected as bullying, it might still be caught as a threat or hate speech. Multilabel models allow detecting all abuse types together. They predict each label independently, thereby increasing the likelihood of detecting harmful behavior. Multilabel detection is better suited to real-world scenarios.It provides an extensive overview of the abuse and facilitates improved decision-making in the moderation system. It also helps platforms take more accurate actions. Recognizing co-occurring abusive categories is important to ensure safer online environments. The growing use of Bangla social media, along with the lack of robust transformer-based multilabel systems, underscores the need for advanced, context-aware NLP models for accurate cyberbullying detection.

Cyberbullying detection is a classic problem in NLP. It is a supervised learning task where predefined categories are assigned to free-text data using linguistic and contextual features. Bangla text classification and cyberbullying detection have used various machine learning (ML), deep learning (DL), and hybrid models to identify abusive content. Traditional ML models include Naïve Bayes, Logistic Regression, Support Vector Machines, Random Forests, Decision Tree, and k-Nearest Neighbors. These models often use handcrafted features, such as n-grams and TF–IDF, for classification tasks, such as cyberbullying detection. Deep learning models such as artificial neural networks, recurrent neural networks, convolutional neural networks, and hybrid DL architectures have also been applied. These models help capture local patterns and long-term dependencies for better representation learning. Many methods still use static word embeddings, such as Word2Vec. These embeddings cannot capture contextual meaning. Transformer-based models solve this by using contextual representations and show better results. Bangla remains under-resourced in NLP. It faces challenges such as complex morphology and lacks robust domain-specific fusion models. This paper presents a framework that addresses these issues for Bangla multilabel cyberbullying detection.

Ml and dl methods have been applied in contrary or in hybrid with other for better classification. Solutions based on various ml like LR, RF, SVM, NB, KNN, DT\cite{balakrishnan2020improving} \cite{islam2020cyberbullying} \cite{balakrishnan2020improving} \cite{singh2022machine} perform well with small data. However, critical challenges still affect their effectiveness. Traditional ML models are commonly used but often fail to interpret the complex language and context of social media posts \cite{Nuthalapati2024CyberbullyingComparison}, \cite{Aljohani2023CyberbullyingReview}. The DL-based solutions, such CNN, RNN, LSTM, BiLSTM, and GRU models \cite{murshed2022dea}\cite{alabdulwahab2023cyberbullying}\cite{nath2024deep}\cite{iwendi2023cyberbullying}, achieved a higher accuracy degree but tended to suffer from the overfitting problem and low generalization performance on the lowresource-learning condition. Hybrid models that perform feature extraction in combination with sequential modeling like CNN-LSTM, CNN-BiLSTM, CNN-GRU, and Transformer-based models \cite{albayari2024cyberbullying}\cite{aldhyani2022cyberbullying}\cite{balaji2024cyberbullying}\cite{bermudez2025measuring} have also raised the state of art performance. Most existing studies on cyberbullying detection focus on single-label classification. These approach cannot handle cases where multiple abuse types appear in one comment. Work on multilabel cyberbullying detection is limited, especially for Bangla text. Existing multilabel studies do not provide generalized frameworks with strong accuracy and robustness. Many models do not fully use context-aware representations, which are essential for understanding abusive intent. Several studies also ignore class imbalance or test only one sampling strategy. Model evaluation is often limited to a single train-test split. Cross-validation is rarely used to verify model stability and generalization. These limitations show the lack of a reliable, context-aware, and well-validated multilabel cyberbullying detection framework for Bangla. These motivate us to develop a robust context-aware hybrid multilabel model that can detect multiple abuse categories simultaneously. Such a framework can support more accurate detection and faster moderation in real-world applications.

To fill the gaps, this paper introduces a robust deep learning approach that combines pretrained Bangla-BERT-Large embeddings with two-layer stacked LSTM networks for Bangla cyberbullying detection. The proposed work has utilized one publicly available cyberbullying dataset. The datasets are evaluated in three forms (imbalanced, undersampled, and oversampled) to have a fair evaluation on different data distributions. To ensure stability, k-fold cross-validation is used, and model evaluation is carried out extensively using multiple evaluation metrics, including Accuracy, Hamming Loss, Precision, Recall, F1-score, MCC, Cohen-Kappa, and ROC curve. Furthermore, it is enhanced with Local Interpretable Model-agnostic Explanations (LIME) to improve the interpretability of model predictions.

\textbf{Major Contributions of This Work}
\begin{itemize}
    \item We have achieved state-of-the-art accuracy on a public multilabel cyberbullying dataset, outperforming all previous works on it.
    \item Based on prior research, we experimented with hybrid models and proposed a fine-tuned Bangla-BERTLarge with 2-layer stacked LSTM fusion for optimal performance.
    \item Unlike prior single-label studies, this work introduces a generalized framework for detecting multiple co-occurring cyberbullying categories.
    \item Five-fold cross-validation and multiple performance metrics are used to demonstrate the proposed model’s generalization, reliability, and fairness.
    \item Model performance is evaluated using accuracy, Hamming Loss, precision, recall, F1-score, MCC, Cohen-Kappa, and ROC, thereby making the framework more reliable and accurate.
    \item An explainable prediction framework is employed using LIME to provide transparent and trustworthy insights into model decisions.
    \item This work advances natural language processing research for low-resource languages and creates a solid baseline for detecting cyberbullying in Bangla.
\end{itemize}

The rest of this article is organized as follows: the related work is summarized in Section \ref{sec:2}. In Section \ref{sec:3}, we describe the proposed methodology in detail, including dataset collection and processing, as well as the training, validation, and test datasets. In Section \ref{sec:4}, the recognition experimental results are presented, along with a detailed explanation of the evaluation criteria for the proposed model. Finally, our conclusion is given in Section \ref{sec:5}.

\section{Literature Review}
\label{sec:2}
Cyberbullying detection in the Bengali language is challenging due to the complex linguistic structure. Bangla text often includes informal expressions, implicit meaning, and mixed writing styles. These features make automatic abuse detection difficult. The task becomes more challenging when multilabel cyberbullying detection is required. A single comment may contain two or more types of abusive behavior at the same time. The model must identify all concurrent abuse categories from the same sentence. These increase the difficulty of detection and classification. These challenges explain why multilabel cyberbullying detection in Bangla remains an open research problem. Previous studies have examined cyberbullying detection. However, the rapid growth of user-generated content shows the need for better and more refined cyberbullying prediction methods to understand abusive behavior. A framework of this kind allows for the simultaneous detection of various concurrent forms of bullying in a given piece of text. Moreover, Bangali cyberbullying multilabel datasets tend to have a high level of class imbalance, which negatively influences the performance of the model. To address this issue, undersampling and oversampling methods are used alongside a hybrid transformer-based robust Bangla-BERT-large with a 2-layer Stacked LSTM framework to improve classification performance. In the existing literature, this combined mechanism for managing class imbalance in multilevel prediction of cyberbullying in Bangla has not been properly studied.

\begin{table*}[htbp]
\centering
\large
\renewcommand{\arraystretch}{1.35}
\resizebox{\textwidth}{!}{%
\begin{tabular}{|
p{1.8cm}|c|c|c|c|c|p{3.0cm}|p{3.8cm}|p{3.1cm}|c|c|c|c|c|c|c|c|c|c|p{4.8cm}|}
\hline

\multirow{2}{*}{\textbf{Study}}
& \multicolumn{5}{c|}{\textbf{Dataset}}
& \multirow{2}{*}{\textbf{ML Models}}
& \multirow{2}{*}{\textbf{DL Models}}
& \multirow{2}{*}{\textbf{Hybrid Models}}
& \multicolumn{10}{c|}{\textbf{Metrics}}
& \multirow{2}{*}{\textbf{Limitations}} \\

\cline{2-6} \cline{10-19}

& \rotatebox{90}{\textbf{English}}
& \rotatebox{90}{\textbf{Bangla}}
& \rotatebox{90}{\textbf{Other}}
& \rotatebox{90}{\textbf{Multi-label}}
& \rotatebox{90}{\textbf{No. of Classes}}

& & &

& \rotatebox{90}{\textbf{Accuracy}}
& \rotatebox{90}{\textbf{Precision}}
& \rotatebox{90}{\textbf{Recall}}
& \rotatebox{90}{\textbf{F1}}
& \rotatebox{90}{\textbf{Hamm. Loss}}
& \rotatebox{90}{\textbf{Kappa}}
& \rotatebox{90}{\textbf{MCC}}
& \rotatebox{90}{\textbf{ROC}}
& \rotatebox{90}{\textbf{Cross Val.}}
& \rotatebox{90}{\textbf{XAI}}
& \\

\hline


\cite{emon2019deep}
& \xmark & \cmark & \xmark & \xmark & 7
& MNB, LR, RF, LinearSVC
& ANN, RNN
& \xmark
& \cmark & \cmark & \cmark & \cmark & \xmark & \xmark & \xmark & \xmark & \xmark & \xmark
& Relatively small dataset and overfitting issues \\ \hline

\cite{roy2022cyberbullying}
& \cmark & \xmark & \xmark & \xmark & 2
& \xmark
& 2DCNN
& \xmark
& \cmark & \cmark & \cmark & \xmark & \xmark & \xmark & \xmark & \cmark & \xmark & \xmark
& Relatively small dataset (1k–3k images), limiting generalization \\ \hline

\cite{bilal2022context}
& \xmark & \xmark & \cmark & \xmark & 2
& LR, SVM, XGB, RF, DT, KNN
& LSTM, CNN, Bi-LSTM
& \xmark
& \cmark & \cmark & \cmark & \cmark & \xmark & \xmark & \xmark & \xmark & \xmark & \xmark
& Uses custom Word2Vec embeddings, lacking transformer-based contextual representations \\ \hline

\cite{almufareh2025integrating}
& \cmark & \xmark & \xmark & \xmark & 6
& LR, RF, XGB, DT, NB, SVM, ET
& \xmark
& \xmark
& \cmark & \cmark & \cmark & \cmark & \xmark & \xmark & \xmark & \cmark & \xmark & \xmark
& TF–IDF-based static features which lack deep contextual and semantic representation. \\ \hline

\cite{akhter2023robust}
& \xmark & \cmark & \xmark & \xmark & 4
& DT, RF, SVM, MLP
& \xmark
& Hybrid ML Approach
& \cmark & \cmark & \cmark & \cmark & \xmark & \xmark & \xmark & \cmark & \cmark & \xmark
& Shallow machine learning models \\ \hline

\cite{mahmud2023cyberbullying}
& \xmark & \cmark & \xmark & \xmark & 2
& LR, DT, RF, MNB, KNN
& BanglaBERT
& \xmark
& \cmark & \cmark & \cmark & \cmark & \xmark & \xmark & \xmark & \cmark & \xmark & \xmark
& Multilabel cyberbullying detection and robust hybrid transformer architectures are not explored \\ \hline

\cite{sifath2024recurrent}
& \xmark & \cmark & \xmark & \xmark & 4
& LR, NB, XGB, RF
& RNN, Tri-RNN
& CNN-LSTM-RNN
& \cmark & \cmark & \cmark & \cmark & \xmark & \xmark & \xmark & \xmark & \xmark & \xmark
& No transformer embeddings explored \\ \hline

\cite{debnath2025multi}
& \xmark & \cmark & \xmark & \xmark & 4
& \xmark
& BanglaBERT, DistilBERT, ElectraBERT, RoBERTa
& \xmark
& \cmark & \cmark & \cmark & \cmark & \xmark & \xmark & \xmark & \xmark & \xmark & \xmark
& Only tested on transformer architectures \\ \hline

\cite{philipo2025assessing}
& \cmark & \xmark & \xmark & \xmark & 5
& \xmark
& BERT, DistilBERT, RoBERTa, XLNet, GPT-2
& \xmark
& \cmark & \cmark & \cmark & \cmark & \xmark & \xmark & \xmark & \xmark & \xmark & \xmark
& Limiting generalization claims \\ \hline

\cite{khan2023multi}
& \cmark & \xmark & \xmark & \cmark & 5
& MNB, RF, LP-LR
& \xmark
& \xmark
& \cmark & \cmark & \cmark & \cmark & \cmark & \xmark & \xmark & \xmark & \xmark & \xmark
& No cross-dataset validation, hence overfitting risks are not deeply analyzed despite multiple label configurations \\ \hline

\cite{islam2023deep}
& \xmark & \cmark & \xmark & \cmark & 5
& RF, LR, MNB, SGD
& Bi-LSTM, LSTM, GRU
& CNN-LSTM
& \cmark & \cmark & \cmark & \cmark & \cmark & \xmark & \xmark & \xmark & \xmark & \xmark
& Based on english language not experiment any low resource language like bangla\\ \hline

\cite{sharif2022m}
& \xmark & \cmark & \xmark & \cmark & 5
& LR
& BiGRU+FT, mBERT, IndicBERT, BanglaBERT
& NB-SVM
& \xmark & \cmark & \cmark & \cmark & \xmark & \xmark & \xmark & \xmark & \xmark & \xmark
& Limited architectural innovation \\ \hline

Our Study
& \xmark & \cmark & \xmark & \cmark & 5
& KNN, NB, LR, XGB, RF
& LSTM, BiLSTM, GRU, BiGRU, BanglaBERT, mBERT, DeBERTa, RoBERTa, indicBERT
& BERT-LSTM, BERT-BiLSTM, BERT-GRU, BERT-BiGRU 
& \cmark & \cmark & \cmark & \cmark & \cmark & \cmark & \cmark & \cmark & \cmark & \cmark
& The framework is restricted to only Bengali language but extension to further multilingual languages are needed for better automatic detection system \\ \hline

\end{tabular}%
}

\small
\textbf{Legend:} \cmark = Yes, \xmark = No, Cross Val. = Cross Validation, Hamm.\ Loss = Hamming Loss, MNB = Multinomial Naive Bayes, LR = Logistic Regression, RF = Random Forest, GB = Gradient Boosting, XGB = Extreme Gradient Boosting, MLP = Multi-Layer Perceptron, DT = Decision Tree, ET = Extra Tree, SGD = Stochastic Gradient Descent, FT = Fast Text, LP-LR = Label Powerset Linear Regression, KNN = K-Nearest Neighbors 
\caption{Comparison of related studies with the proposed BanglaBERT-LSTM hybrid model.}
\label{tab:comparision_table}
\end{table*}

Some earlier works on automatic detection of singlelabel Bangla cyberbullying highlighted the challenges involved in this area, primarily because of the unavailability of datasets and accurate word embeddings. An early endeavor found that RNN outperforms classical ML classifiers, i.e., Multinomial Naive Bayes, LinearSVC, Logistic Regression, and Random Forest, on TF-IDF and CountVectorizer for a self-collected 4700 Bangla cyberbullying comments\cite{emon2019deep}. Based on this understanding, the later studies attempted to study deep learning and hybrid methods as well. For example, a two-dimensional CNN-based deep learning model was tested on two self-collected datasets of 1000 and 3000 cyberbullying texts collected from images by using transfer learning under different optimizers and learning rate variations \cite{roy2022cyberbullying}. Another work also implemented a low-resource language like Urdu using various classical ML classifiers, i.e., SVM, Naïve Bayes, Logistic Regression, XGBoost, Decision Tree, and KNN, along with a neural network, achieving 87.50\% accuracy as the best performance by the Bi-LSTM with an additional attention layer among all \cite{bilal2022context}. Additionally, a researcher analyzed all ML models utilizing SMOTE approaches for addressing class imbalance and discovered that the Extra Tree models surpassed all other models with an accuracy of 95\% \cite{almufareh2025integrating}. One hybrid ML approach with InstanceHardnessThreshold (IHT) has been tested in one publicly available English cyberbullying dataset, and the authors show that this hybrid model attained an accuracy of 98\% across five different cyberbullying categories \cite{akhter2023robust}. In addition, Bangla-BERT outperformed numerous ML models with different feature extraction algorithms, achieving an accuracy of 94\% while being able to maintain precision, recall, and F1 above 93\% \cite{mahmud2023cyberbullying}. In the same way, more complicated designs have been proposed; an example of such is a hybrid CNN+LSTM+RNN fusion model that achieved 86\% accuracy\cite{sifath2024recurrent}. Later, the transformer-based model has also been proposed; an example is where the author found that Bangla-BERT reported the highest accuracy of 90\% with a 91\% F1-score over all other transformer models such as BERT, ElectraBERT, RoBERTa, DistilBERT, and MBERT on the Bengali language\cite{debnath2025multi}. Another research work on English cyberbullying datatsets has found that RoBERTa outperforms other transformer-based models with 96\% accuracy, whereas the second-best model, BERT, reported 95\% accuracy\cite{philipo2025assessing}. After focusing on single-label cyberbullying detection, few researchers explore multilabel cyberbullying detection. Traditional ML models have been tested on 4 categorical combinations where the Label Powerset-Logistic Regression obtains 88\%, 90\%, 91\%, and 93\% accuracy accordingly for 5-label, 4-label, 3-label, and 2-label cyberbullying datasets datasets\cite{khan2023multi}. Similarly, the authors of \cite{islam2023deep} conducted numerous studies to detect multilevel cyberbullying on a publicly available english multilabel datasets with 5 distinct classes. They explored various DL architectures, such as LSTM, BiLSTM, and BiGRU, to develop advanced cyberbully detection systems. They found one CNN-LSTM hybrid model that outperforms all other models with the accuracy and F1-score of 87.8\% and 88.3\%\cite{islam2023deep}. A novel multilabel Bengali dataset (named MBAD) containing 15650 texts to detect aggressive texts and their targets was introduced, where BanglaBERT acquired the highest weighted F1 score in both detection (0.92\%) and target identification (0.83\%) tasks \cite{sharif2022m}.

Although previous research contributed to the detection of Bangla cyberbullying, it possesses several limitations. multilevel is one of them, all them largely focus on single level, there are hardly any work which has not been proposed so far, only which is existed over only bengali multilevel dataset which achieved 93\% where we have proposed a more robust version of it from various experiments. In contrast to contextualized transformer embeddings, most methods rely on static feature representations like TF–IDF and CountVectorizer, which lack contextual awareness and are unable to capture subtle semantic links. Most previous studies use single-label classification, which ignores the co-occurrence of several cyberbullying categories in a single incident. Additionally, there hasn't been enough research done on reliable hybrid architectures that combine transformers with complementary learning processes. Many studies use short or unbalanced datasets, are limited to binary or low-granularity class settings, and lack sufficient preparation methods for Bangla, such as stopword coverage, lemmatization, and abbreviation normalization. Furthermore, cross-validation techniques are frequently lacking or used inconsistently, which raises questions regarding the generality and durability of the results. Additionally, existing hybrid methods frequently exhibit overfitting due to shallow feature fusion, while reliance on either shallow machine learning models or isolated deep learning frameworks limits their capacity to model the linguistic and contextual complexity of Bangla text.

In response to these limitations, we have proposed a novel transformer based robust hybrid deep learning framework, BERT-LSTM, for Bangla cyberbullying. Our model fuses transformer-based contextual embedding (BERT), and two sequential dependency modelling (LSTM) to combine local and global semantics in a transformer-based approach for cyberbullying texts. In these studies, the publicly available "Bangla multilabel cyberbully, sexual harassment, threat, and spam detection dataset" imbalanced dataset that has not been explored so far in this context is used. Class distribution is balanced using undersampling and oversampling methods for bias reduction and better model performance. In this study, the hybrid model BERT-LSTM outperformes the previous state of state on this datasets.

\begin{figure*}[t] 
    \centering
    \includegraphics[width=\linewidth]{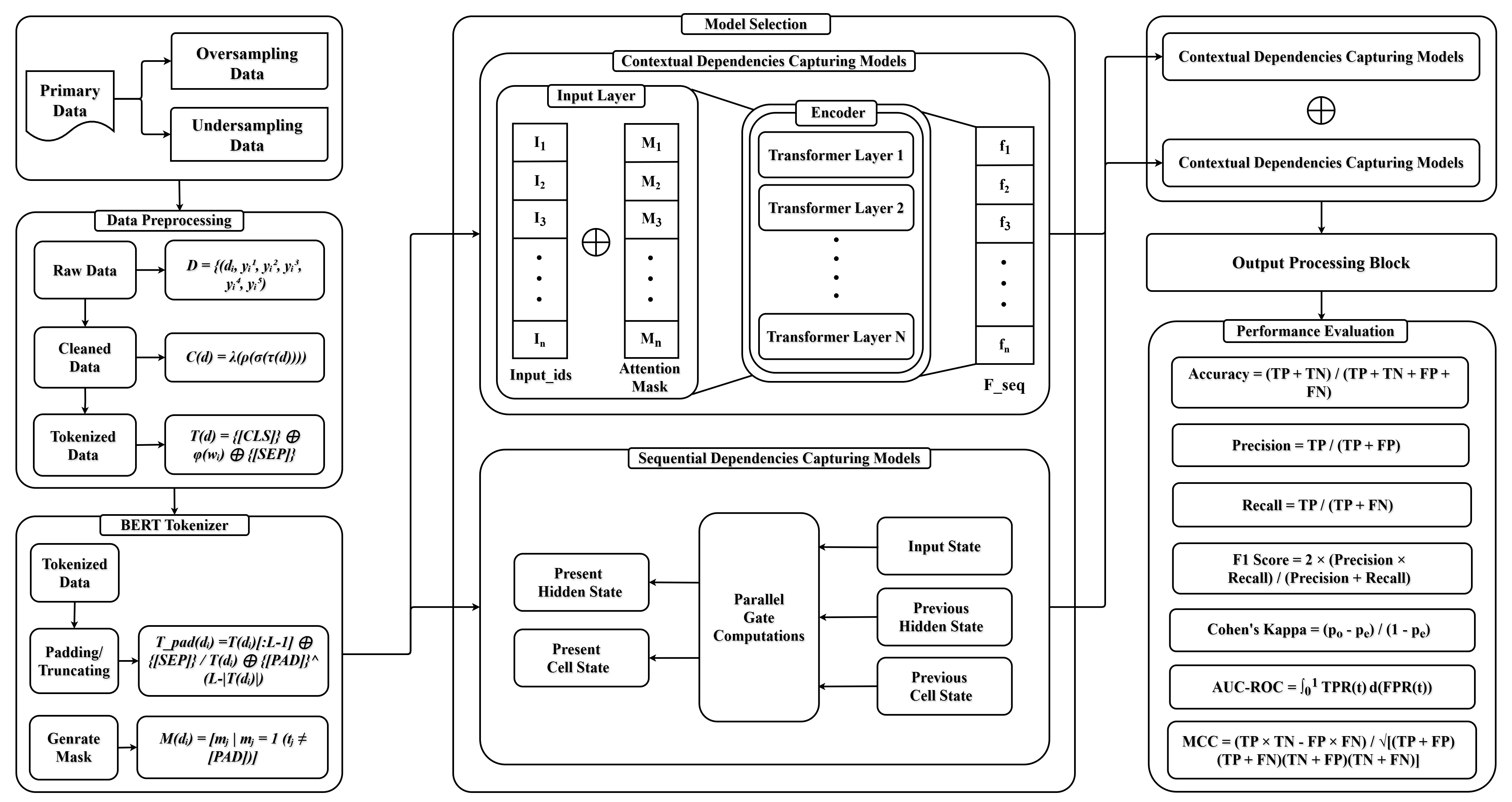}
    \caption{Workflow of the Proposed Cyberbullying Detection}
    \label{fig-workflow}
\end{figure*}

\section{Methodology}
\label{sec:3}
We have designed a framework in this study to address the issue of creating multilevel cyberbullying detection in layers of preprocessing, embedding features into context, and sequential feature extraction. This begins with collecting raw data, after which oversampling and undersampling are employed to address class imbalance. Then the data passes through a complete preprocessing pipeline, which cleans up the text, removes stopwords, special characters, and performs tokenization, etc. Then, the sequences are encoded with a BERT tokenizer using padding or truncation, and attention masks are created to ensure all inputs are of equal length. The resulting representations are sent to a pretrained BERT-Large model. Deep representations on a sentence-level are obtained due to the content-rich semantics of the transformer layers of BERT, which extract word-wise and contextual semantics. In order to introduce extra richness, we feed those BERT embeddings into a 2-layer stacked LSTM, which captures long-term sequential dependencies that BERT may fail to capture. The dropout layers help prevent overfitting, and the dense and activation layers generate a discriminative feature vector for the final classification. The combination of the transformer's power and the LSTM's time modeling results in a more detailed feature space.

The entire workflow is described in Figure \ref{fig-workflow}, which provides a summary of the most crucial steps of data preparation, feature extraction, and classification. 

\subsection{Dataset Collection}
The dataset is the most vital factor in this research. Most of the time spent on this research is choosing and checking the dataset's quality. The dataset used in this research is a combination of datasets collected from \cite{Sunny2024BanglaMultilabelDataset}, serving as the foundational corpus for multilevel cyberbullying detection. The dataset consists of 12,557 Bangali social media comments collected from Facebook and TikTok. It is specifically preselected to make it easier to learn to identify harmful and abusive online behaviour automatically. All comments are marked on five binary types of abuse, including Bully, Sexual, Religious, Threat, and Spam, and thus make it possible to have a multi-label representation where a comment can have multiple forms of abusive behavior. The sample was obtained from a preliminary set of about 30,000 comments containing Bangla, English, and Banglish (Bangla in the Latin script). During sample preparation, all non-Bangla records were removed, followed by the removal of duplicate, null, and noisy records. Additional text cleaning was performed to remove residual English characters and numerical symbols. Formally, the dataset can be expressed as
\[
D = \{(x_i, y_i^{\text{bully}}, y_i^{\text{sexual}}, y_i^{\text{religious}}, y_i^{\text{threat}}, y_i^{\text{spam}}) \mid i = 1, \ldots, N\},
\]
where \(x_i\) denotes the \(i\)-th comment and each \(y_i^{(\cdot)} \in \{0,1\}\) indicates the presence or absence of the corresponding abuse category. Inherent class imbalance exists in the corpus, with some abuse categories being represented at much higher frequencies. This imbalance motivates the development and evaluation of the multilevel structure for cyberbullying detection presented here.

\begin{table*}[htbp]
\centering
\small
\renewcommand{\arraystretch}{0.50}
\resizebox{\textwidth}{!}{%
\begin{tabular}{|l|l|c|c|c|c|c|}
\hline
\textbf{Dataset Split} & \textbf{Dataset Type} & \multicolumn{5}{c|}{\textbf{Class}} \\
\cline{3-7}
& & \textbf{Bully} & \textbf{Sexual} & \textbf{Religious} & \textbf{Threat} & \textbf{Spam} \\
\hline
\multirow{3}{*}{\textbf{Train}} 
& Imbalance & 5497 & 1447 & 1140 & 1116 & 752 \\
\cline{2-7}
& Oversample & 16472 & 6189 & 7418 & 6841 & 5986 \\
\cline{2-7}
& Downsample & 2685 & 1047 & 1163 & 1142 & 835 \\
\hline
\textbf{Validation} 
& - & 643 & 105 & 134 & 145 & 77 \\
\hline
\textbf{Test}
& - & 643 & 105 & 134 & 145 & 77 \\ 
\hline
\end{tabular}%
}
\caption{Dataset Statistics and Class Distribution}
\label{tab:dataset_stats}
\end{table*}

\subsection{Undersampling and Oversampling}
The data set is heavily imbalanced with respect to the five different abuse classes, showing significant disparity in the frequencies of different abusive behaviors. Therefore, undersampling and oversampling techniques are applied only to the training split to ensure reliable and effective model training. This decision maintains the original distribution of the validation and test sets and stops data leakage. The sampling process follows multilabel behavior rather than strict single-label rules. In a multilabel dataset, a single comment can belong to multiple abuse categories simultaneously. When a minority class is oversampled, the same comment and all of its labels are repeated.  This duplication can also increase the number of samples in the majority class, such as the bullying class. This behavior is expected and does not indicate an error in the sampling process. The class distributions for the original, undersampled, and oversampled training sets are shown in Table~\ref{tab:dataset_stats}.

Let \(n_c\) denote the number of instances belonging to abuse category \(c\). The resampling procedures modify the class distributions as follows.

\[
n_c^{\text{under}} = \min(n_c).
\]
The resulting undersampled dataset is
\[
D^{\text{under}} = 
\left\{
\begin{array}{l}
(x_i,\ y_i^{\text{bully}},\ y_i^{\text{sexual}},\ y_i^{\text{religious}}, \\
\quad y_i^{\text{threat}},\ y_i^{\text{spam}}) 
\mid i = 1,\dots, N
\end{array}
\right\},
\quad N = n_c^{\text{under}}.
\]

\[
n_c^{\text{over}} = \max(n_c).
\]
The oversampled dataset is defined as
\[
D^{\text{over}} = 
\left\{
\begin{array}{l}
(x_i,\ y_i^{\text{bully}},\ y_i^{\text{sexual}},\ y_i^{\text{religious}}, \\
\quad y_i^{\text{threat}},\ y_i^{\text{spam}}) 
\mid i = 1,\dots, N
\end{array}
\right\},
\quad N = n_c^{\text{over}}.
\]

The resampling process ensures more balanced representation of abuse categories in the undersampled and oversampled training data. This balance reduces bias during model training and improves the robustness of cyberbullying prediction.

\subsection{Data Preprocessing}
This research uses a full preprocessing process, which involves data cleaning, stemming, tokenization and padding with the use of token level encoding. These are operations that takes in raw comment text and output a structured, embeddable, and machine-interpretable receipt suitable for end classification. The formal definition of the preprocessing workflow is presented below with respect to each step and its mathematical formulation. 

\begin{itemize}
\item\textbf{Clean Data:} Several forms of noise are presented in the raw multilevel cyberbullying dataset, including non-Bangla alphabets, repeated characters and symbols, numerals, and extraneous metadata. In order to obtain a streamlined and analytically useful corpus, such unnecessary elements are removed, and the textual content is normalised. This cleaning procedure ensures that each comment is stripped of irrelevant material while preserving its semantic integrity. 

Let the cleaned dataset be denoted by \( C(d) \), defined through the composite transformation
\[
C(d) = \lambda\bigl( \rho( \sigma( \tau(d) ) ) \bigr),
\]
where \( \tau(\cdot) \) performs text normalisation, \( \sigma(\cdot) \) removes noise and irrelevant characters, \( \rho(\cdot) \) eliminates stopwords, and \( \lambda(\cdot) \) applies the final lexical refinement.

\item \textbf{Tokenization:} 
Tokenization breaks cleaned and processed text into smaller units called tokens. These tokens are later converted to numerical representations for deep learning models. In Bangla, tokenization captures complex word forms and meanings using subword segmentation. For each comment $d_i$, each word $w_j$ is mapped to its subword representation using a function $\phi(w_j)$. Special tokens [CLS] \& [SEP] are added to mark the start and end of the sequence. This format helps the model understand sentence boundaries and structure. 

Let the token be denoted by \( T(d_i) \) is defined by,
\[
T(d_i) = \{[\text{CLS}]\} \,\oplus\, \phi(w_j) \,\oplus\, \{[\text{SEP}]\},
\]
where \( \oplus \) denotes sequence concatenation. This programmed token sequence then forms the basis for subsequent padding, encoding, and embedding within the model. 
\end{itemize}

\subsection{BERT Tokenizer}
The BERT tokenizer performs two essential preprocessing operations that prepare each token sequence for transformer-based encoding, because BERT requires all input sequences to have a fixed length \(L\), the tokenizer standardizes variable-length comments while maintaining the semantic and structural integrity of the original text. The two core components of this tokenizer are described below:

\paragraph{\textbf{Padding and Truncation.}}
For each tokenized sequence \(T(d_i)\), if its length exceeds the maximum allowed length \(L\), the sequence is truncated; if it is shorter, padding tokens \([\text{PAD}]\) are appended. Formally,
\[
    T_{\text{pad}}(d_i) =
    \begin{cases}
    T(d_i)[L{-}1] \;\oplus\; \{[\text{SEP}]\}, & \text{if } |T(d_i)| > L, \\[6pt]
    T(d_i) \;\oplus\; [\text{PAD}]^{\,(L - |T(d_i)|)}, & \text{if } |T(d_i)| < L.
    \end{cases}
    \]
    This ensures that every input sequence has the fixed length required by the BERT encoder.

    \paragraph{\textbf{Attention Mask Generation.}}
    The attention mask discriminates between real tokens and padded positions, allowing the transformer to ignore padding during self-attention computation. A mask value of \(1\) signifies a valid token, whereas \(0\) indicates padding. For each mask value m, the mask is defined as
    \[
    M(d_i) = \{ m_j \mid 
    m_j =
    \begin{cases}
    1, & \text{if } t_j \neq [\text{PAD}], \\[4pt]
    0, & \text{otherwise},
    \end{cases}
    \}
    \]
    where \( t_j \) denotes the \( j\text{-th} \) token in the padded sequence.

\begin{figure*}[t] 
    \centering
    \includegraphics[width=\linewidth]{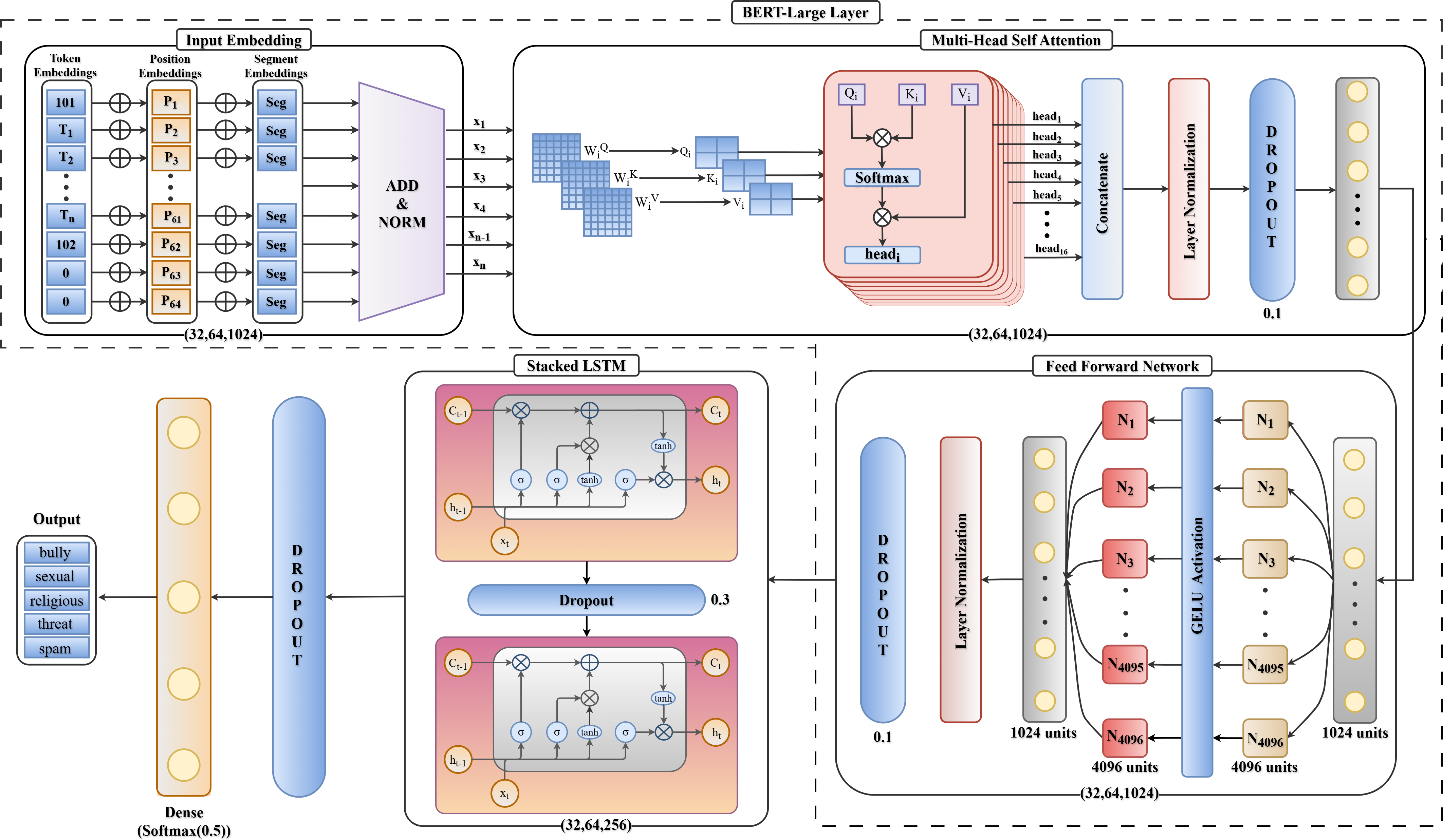}
    \caption{Model Architecture of Proposed BERT-CNN-BiLSTM}
    \label{fig-model_archi_bert-lstm}
\end{figure*}

\subsection{Proposed Model Architecture Overview}
The overall structure of the proposed model in the suggested hybrid process is shown in Fig \ref{fig-model_archi_bert-lstm}. This model begins with the BERT-Large embedding layer that produces contextual representations of its embedding and transformer blocks. The representations of the representations containing the temporal component and long-range dependencies are then encoded with two consecutive LSTM layers via stacked bi-directional LSTM module. The dropout regularizes the features and passes them through a dense layer to determine the multi-level categories of abuse. The parts of the architecture are then described in detail, beginning with the Bert embedding layer.

\subsubsection{\textbf{BanglaBERT-Large Layer}}
BanglaBERT-Large is the primary feature extractor in our system. It provides context-aware vector representations from unprocessed Bengali text. Both surface-level word patterns and more profound semantic meaning are captured. BanglaBERT-Large is a bidirectional Transformer encoder. The entire input sequence is processed all at once. This enables it to understand each word's meaning and how it relates to other words. Such contextual understanding is important for classification. The meaning of a sentence can change based on nearby words. A small word or phrase can make a sentence offensive. BanglaBERT helps to capture this subtle change in meaning. For our BERT-Large layer, the input parameters we used are:  Maximum sequence length which is denoted by $max\_len$ is 64 tokens, Hidden dimension $d_{\text{bert}}$ is 1024, Batch size $B$ is 32, Token IDs $I \in \mathbb{R}^{32 \times 64}$, Attention masks $M \in \{0,1\}^{32 \times 64}$ indicating valid tokens.

\paragraph{\textbf{Input Embedding Layer}}  
Before passing the Bengali text to the BanglaBERT-Large encoder, each token is converted into a dense vector. This vector is made up of three parts: token embedding, positional embedding, and segment embedding. Let the input sequence be  $I = [I_1, I_2, \ldots, I_n],$ where $n$ is the number of tokens in the input, and each $I_i$ represents the $i^{\text{th}}$ token. Each token $I_i$ is mapped to a token embedding using the matrix  $W \in \mathbb{R}^{V \times d_{\text{bert}}}$, where $V$ is the size of the vocabulary and $d_{\text{bert}}$ is the embedding dimension (set to 1024 in BanglaBERT-Large). This captures the semantic meaning of the token. Positional information is added using a positional embedding matrix $P \in \mathbb{R}^{\text{max\_len} \times d_{\text{bert}}}$, which assigns a unique vector to each position in the sequence to help the model understand word order. Segment identity is added using segment embeddings $E_{\text{seg}}$, which distinguish different parts of the input (e.g., sentence A vs. sentence B in next sentence prediction tasks). The final embedding for each token is the sum of these three components. The combined embedding for token $i$ is therefore defined as  
\[
h_i^{(0)} = e_{\text{token}}(i) + e_{\text{pos}}(i) + e_{\text{seg}}(i) \in \mathbb{R}^{d_{\text{bert}}},
\]  
where \( e_{\text{token}}(i) \) is the token embedding for \( I_i \), \( e_{\text{pos}}(i) \) is the positional embedding at position \( i \), \( e_{\text{seg}}(i) \) is the segment embedding for the segment that \( I_i \) belongs to.

Layer Normalization and dropout are then applied to this combined vector to stabilize training and reduce overfitting. The output becomes:  
\[
x_i = \text{Dropout}\left(\text{LayerNorm}\left(h_i^{(0)}\right),\, p=0.1\right),
\]  
where \( p = 0.1 \) is the dropout rate.

For the entire input sequence, the embedding layer constructs the following matrix of token representations:  
\[
X^{(0)} = [\,x_1, x_2, \ldots, x_n\,] \in \mathbb{R}^{n \times d_{\text{bert}}},
\]  
which serves as the standardized input to the BanglaBERT-Large Transformer encoder.

\paragraph{\textbf{Transformer Encoder Layers}  }
BanglaBERT-Large employs a stack of 24 Transformer encoder layers, each consisting of a multi-head self-attention (MHSA) module and a position-wise feed-forward network (FFN), wrapped with residual connections, dropout, and layer normalization. Given the input to layer \(l\),
\[
H^{(l-1)} = [h_1^{(l-1)}, h_2^{(l-1)}, \ldots, h_n^{(l-1)}] \in \mathbb{R}^{n \times d_{\text{bert}}},
\]
the encoder progressively refines contextual representations by integrating both local and global linguistic cues relevant to cyberbullying detection.

    \noindent\textbf{Multi-Head Self-Attention (MHSA)}  
    For each token representation \(h_i^{(l-1)}\), the query, key, and value projections are computed as per \cite{vaswani2017attention}
    \[
    \mathbf{q}_i^{(l)} = \mathbf{W}_Q^{(l)} h_i^{(l-1)}, \quad
    \mathbf{k}_i^{(l)} = \mathbf{W}_K^{(l)} h_i^{(l-1)}, \quad
    \mathbf{v}_i^{(l)} = \mathbf{W}_V^{(l)} h_i^{(l-1)}, 
    \]
    where the query, key, value weight matrices \(\mathbf{W}_Q^{(l)}, \mathbf{W}_K^{(l)}, \mathbf{W}_V^{(l)} \in \mathbb{R}^{d_{\text{bert}} \times d_{\text{bert}}}\). 
    These projections are partitioned into \(h = 16\) heads, each with sub-dimension \(d_{\text{sub}} = d_{\text{bert}}/h = 64\). Scaled dot-product attention is also computed as per \cite{vaswani2017attention}
    \[
    \mathrm{Attention}(Q^{(l)}, K^{(l)}, V^{(l)})
    = \mathrm{softmax}\!\left(\frac{Q^{(l)} (K^{(l)})^{T}}{\sqrt{d_k}}\right) V^{(l)},
    \]
    producing attention maps \(A^{(l)} \in \mathbb{R}^{n \times n}\).  
    The Outputs of all heads are concatenated and linearly transformed\cite{vaswani2017attention}:
    \[
    \mathrm{MultiHead}(H^{(l-1)}) 
    = \mathrm{Concat}(\mathrm{head}_1, \ldots, \mathrm{head}_{16}) W_O^{(l)},
    \]
    where \(W_O^{(l)} \in \mathbb{R}^{d_{\text{bert}} \times d_{\text{bert}}}\).  
    This allows the encoder to capture diverse cyberbullying cues.

    \noindent\textbf{Add \& Norm}  
    Residual connections stabilize learning in deep architectures. With sub-layer output \(f(i)\), the intermediate residual output,
    \[
    r^{(l)}(i) = h^{(l-1)}(i) + \mathrm{Dropout}(f(i)),\]
    The hidden input representation of token \(i\) from the previous layer \((l-1)\), denoted as \(h^{(l-1)}(i)\),
    \[
    h_{\text{norm}}^{(l)}(i) = \mathrm{LayerNorm}(r^{(l)}(i)).
    \]
    
    \noindent\textbf{Feed-Forward Network (FFN)}  
    Following the position-wise feed-forward network defined in the original Transformer architecture \cite{vaswani2017attention}, each BanglaBERT encoder layer applies a two-layer fully connected network with GELU activation, as adopted in BERT \cite{devlin2019bert}:
    \[
    \mathrm{FFN}(i) 
    = \mathrm{GELU}\!\left( h_{\text{norm}}^{(l)}(i) W_1^{(l)} + b_1^{(l)} \right) W_2^{(l)} + b_2^{(l)},
    \]
    where \(W_1^{(l)} \in \mathbb{R}^{d_{1024} \times d_{\text{ff}}}\) with \(d_{\text{ff}} = 4096\),  
    and \(W_2^{(l)} \in \mathbb{R}^{4096 \times 1024}\).  
    The output is refined via another normalization step:
    \[
    \begin{aligned}
    r_{\text{ffn}}^{(l)}(i) &= h_{\text{norm}}^{(l)}(i) + \mathrm{Dropout}(\mathrm{FFN}(i)), \\
    H^{(l)}(i) &= \mathrm{LayerNorm}(r_{\text{ffn}}^{(l)}(i)).
    \end{aligned}
    \]
    
The encoder forms a deep linguistic hierarchy. The initial layers (1-6) capture morphological features, part-of-speech cues, and shallow abusive expressions. The middle layers (7-18) focus on sentence-level meaning, including sarcasm, polarity, and the identification of abuse targets. The higher layers (19-24) model discourse-level intent, severity, and more abstract forms of abusive behavior. Together, these layers generate multi-level contextual embeddings that are essential for detecting complex forms of cyberbullying in Bengali text.

\begin{algorithm}[t]
\caption{Feature Extraction using BERT Embeddings}
\label{alg:bert_internal}
\begin{algorithmic}[1]

\Require Sentence $S$, tokenizer $\text{TokBERT}$, BERT encoder with $N_{\text{layers}}=24$, max sequence length $L_{\max}=64$, hidden size $d_{\text{bert}}=1024$, heads $n=16$
\Ensure BERT embeddings $H_{\text{drop}} \in \mathbb{R}^{B \times L_{\max} \times d_{\text{bert}}}$

\State \textbf{Tokenization and Input Preparation}
\For{$b = 1$ to $B$}
    \State $\text{tokens}[b] \gets [\text{CLS}] + \text{TokBERT}(S[b]) + [\text{SEP}]$
    \State $\text{tokens}[b] \gets$ pad/truncate to $L_{\max}$
    \State $\text{input\_ids}[b] \gets \text{convert\_to\_ids}(\text{tokens}[b])$
    \State $\text{attention\_mask}[b] \gets [1 \text{ if token}\neq[\text{PAD}] \text{ else } 0]$
\EndFor

\State \textbf{Embedding Layer}
\State $E \gets \text{TokenEmbed}(\text{input\_ids}) + \text{PosEmbed}(0\!:\!L_{\max}) + \text{SegmentEmbed}(\text{segment\_ids})$
\State $h^{(0)} \gets \text{LayerNorm}(E)$

\State \textbf{Transformer Encoder Layers}
\For{$\ell = 1$ to $N_{\text{layers}}$}
    \State \textbf{Multi-Head Self-Attention}
    \For{$\text{head}=1$ to $n$}
        \State $d_k \gets d_{\text{bert}}/n$
        \State $Q^{(\text{head})} \gets h^{(\ell-1)} W_Q^{(\text{head})}$
        \State $K^{(\text{head})} \gets h^{(\ell-1)} W_K^{(\text{head})}$
        \State $V^{(\text{head})} \gets h^{(\ell-1)} W_V^{(\text{head})}$
        \State $\text{scores}^{(\text{head})} \gets (Q^{(\text{head})}(K^{(\text{head})})^T)/\sqrt{d_k}$
        \For{$(i,j)$ where $\text{attention\_mask}[j] = 0$}
            \State $\text{scores}^{(\text{head})}[i,j] \gets -\infty$
        \EndFor
        \State $\text{attn\_weights}^{(\text{head})} \gets \text{softmax}(\text{scores}^{(\text{head})})$
        \State $\text{attn\_output}^{(\text{head})} \gets \text{attn\_weights}^{(\text{head})} V^{(\text{head})}$
    \EndFor

    \State $\text{MultiHead} \gets \text{Concat}(\text{attn\_output}^{(1:h)}) W_O$
    \State $H_{\text{norm}}^{(\ell)} \gets \text{LayerNorm}\left(h^{(\ell-1)} + \text{Droupout(MultiHead)}\right)$

    \State \textbf{Feed-Forward Network}
    \State $\text{FFN} \gets \text{GELU}(H_{\text{attn}}^{(\ell)}W_1 + b_1)W_2 + b_2$
    \State $H^{(\ell)} \gets \text{LayerNorm}(H_{\text{attn}}^{(\ell)} + \text{Droupout(FFN)})$
\EndFor

\State $H_{\text{drop}} \gets H^{(N_{\text{layers}})}$
\State \Return $H_{\text{drop}}$

\end{algorithmic}
\end{algorithm}

\begin{algorithm}[t]
\caption{LSTM Sequence Processing}
\label{alg:lstm_processing}
\begin{algorithmic}[1]

\Require BERT embeddings \(H_{\text{drop}} \in \mathbb{R}^{B \times L \times d_{\text{bert}}}\);
hidden size \(d_h\); LSTM layers \(n_{\text{layers}}=2\);
dropout probability \(p=0.3\); batch size \(B=32\)

\Ensure Final hidden state vector
\(h_{\text{final}} \in \mathbb{R}^{B \times d_h}\)

\vspace{0.2cm}
\State Initialize hidden and cell states:
\(h_0^{(l)}=0,\; C_0^{(l)}=0\)

\For{$b=1$ to $B$}
    \For{$l=1$ to $n_{\text{layers}}$}
        \For{$t=1$ to $L$}

            \If{$l=1$}
                \State \(x_t \gets H_{\text{drop}}[b,t]\)
            \Else
                \State \(x_t \gets h_t^{(l-1)}\)
            \EndIf

            \State Concatenate input:
            \State \(z_t \gets [h_{t-1}^{(l)},x_t]\)

            \vspace{0.15cm}
            \State \textbf{Forget Gate}
            \State \(f_t^{(l)} \gets \sigma(W_f^{(l)} z_t + b_f^{(l)})\)

            \State \textbf{Input Gate}
            \State \(i_t^{(l)} \gets \sigma(W_i^{(l)} z_t + b_i^{(l)})\)

            \State \textbf{Candidate Cell Memory}
            \State \(\tilde{C}_t^{(l)} \gets \tanh(W_c^{(l)} z_t + b_c^{(l)})\)

            \State \textbf{Cell State Update}
            \State \(C_t^{(l)} \gets f_t^{(l)} \odot C_{t-1}^{(l)}
            + i_t^{(l)} \odot \tilde{C}_t^{(l)}\)

            \State \textbf{Output Gate}
            \State \(o_t^{(l)} \gets \sigma(W_o^{(l)} z_t + b_o^{(l)})\)

            \State \textbf{Hidden State Output}
            \State \(h_t^{(l)} \gets o_t^{(l)} \odot \tanh(C_t^{(l)})\)

            \vspace{0.15cm}
            \If{$l < n_{\text{layers}}$}
                \State Apply dropout:
                \State \(h_t^{(l)} \gets \text{Dropout}_p(h_t^{(l)})\)
            \EndIf

        \EndFor
    \EndFor

    \State Final representation:
    \State \(h_{\text{final}}[b] \gets h_L^{(n_{\text{layers}})}\)

\EndFor

\State \Return \(h_{\text{final}}\)

\end{algorithmic}
\end{algorithm}

\begin{algorithm}[t]
\caption{Training Procedure of the BERT–LSTM Hybrid Classifier with BCE Loss}
\label{alg:bert_lstm_training}
\begin{algorithmic}[1]

\Require Sentences $S$, labels $Y\in\{0,1\}^{B\times K}$,
learning rate $\alpha$, epochs $E$,
dropout $p=0.3$, clipping threshold $\tau$

\Ensure Training loss $L$ or predicted labels $\tilde{y}$

\State Initialize BERT encoder (Alg.~\ref{alg:bert_internal})
\State Initialize stacked LSTM (Alg.~\ref{alg:lstm_processing})
\State Initialize classifier weights $(W_{\text{clf}},b_{\text{clf}})$
\State Initialize AdamW optimizer and warmup scheduler

\vspace{0.2cm}
\If{training mode}

\For{epoch $=1$ to $E$}
\For{each minibatch $(S_{\text{batch}},Y_{\text{batch}})$}

\State $H_{\text{drop}} \gets$ Alg.~\ref{alg:bert_internal}$(S_{\text{batch}})$
\State $h_{\text{final}} \gets$ Alg.~\ref{alg:lstm_processing}$(H_{\text{drop}})$

\State \textbf{Dense Projection with Dropout:}
\State $z \gets W_{\text{clf}}\cdot\mathrm{Dropout}(h_{\text{final}},p)+b_{\text{clf}}$

\State \textbf{Sigmoid Multi-Label Probabilities:}
\For{$k=1$ to $K$}
\State $\hat{y}_k \gets \sigma(z_k)=\dfrac{1}{1+e^{-z_k}}$
\EndFor

\State \textbf{Binary Cross-Entropy Loss:}
\State
\[
\begin{aligned}    
    L \gets -\frac{1}{B}\sum_{i=1}^{B}\sum_{k=1}^{K}
    \Big[
        &y_{i,k}\log(\hat{y}_{i,k})\\
        &+(1-y_{i,k})\log(1-\hat{y}_{i,k})
    \Big]
\end{aligned}
\]

\State \textbf{Backpropagate gradients:} 
\State \(\nabla_{\theta}L \gets \mathrm{Backward}(L)\)

\State \textbf{Gradient Clipping:}
\If{$\|\nabla_{\theta}L\|_2>\tau$}
\State $\nabla_{\theta}L \gets \tau\dfrac{\nabla_{\theta}L}{\|\nabla_{\theta}L\|_2}$
\EndIf

\State Update parameters: optimizer.step()
\State Update learning rate: scheduler.step()
\State Reset gradients: optimizer.zero\_grad()

\EndFor
\EndFor

\State \Return $L$

\Else

\State \textbf{Inference Mode:}
\State Compute $\hat{y}=\sigma(z)$
\State Predict labels $\tilde{y}_k=(\hat{y}_k\ge0.5)$
\State \Return $\tilde{y},\hat{y}$

\EndIf

\end{algorithmic}
\end{algorithm}

\subsubsection{\textbf{LSTM Layers}}
BanglaBERT-Large provides rich, context-aware token representations. Nevertheless, Transformer encoders do not explicitly model word order. Sequential memory is absent from them. As a result, we combine the BERT layer with a two-layer stacked LSTM. The LSTM captures the abusive or harmful meaning that develops throughout a Bangla sentence. The contextual features are contained in BanglaBERT-Large's final output. These features are then passed to the LSTM for learning sequential patterns. This output is defined as
\[
H^{(24)} = [H^{(24)}(1), \ldots, H^{(24)}(n)], \qquad H^{(24)}(t) \in \mathbb{R}^{1024},
\]
where \(n\) denotes the sequence length and each vector encodes contextual information for a single token. The sequence is then passed to the stacked LSTM layers for further processing. Each LSTM layer analyzes the input step by step over time. This process enables the model to learn both short-term patterns and long-range dependencies within the sequence.

\paragraph{\textbf{LSTM Gate Computation}}   
In a Long Short-Term Memory (LSTM) network, the flow of information is regulated through a set of gating mechanisms that are computed in parallel at each time step. Given the current input vector \(x_t\) and the previous hidden state \(h_{t-1}\), the LSTM first computes the forget gate, which determines how much of the previous cell memory \(C_{t-1}\) should be retained\cite{ahmadi2021lstmfacility}:

\[
f_t = \sigma \left(W_f \cdot [h_{t-1}, x_t] + b_f \right).
\]

Next, the input gate decides which new information should be written into the cell state\cite{ahmadi2021lstmfacility}:

\[
i_t = \sigma \left(W_i \cdot [h_{t-1}, x_t] + b_i \right),
\]

while a candidate memory vector is generated using a hyperbolic tangent activation function\cite{ahmadi2021lstmfacility}:

\[
\tilde{C}_t = \tanh \left(W_c \cdot [h_{t-1}, x_t] + b_c \right).
\]

The updated cell state \(C_t\) is then computed by combining the retained past information and the newly selected candidate content\cite{ahmadi2021lstmfacility}:

\[
C_t = f_t * C_{t-1} + i_t * \tilde{C}_t.
\]

After updating the memory, the output gate controls which parts of the cell state should be exposed as the hidden representation\cite{ahmadi2021lstmfacility}:

\[
o_t = \sigma \left(W_o \cdot [h_{t-1}, x_t] + b_o \right).
\]

Finally, the hidden state output at time step \(t\) is obtained by applying a tanh activation to the cell state and modulating it through the output gate\cite{ahmadi2021lstmfacility}:

\[
h_t = o_t * \tanh(C_t).
\]

Here, \(\sigma(\cdot)\) denotes the sigmoid activation function, \(W_f, W_i, W_c, W_o\) are learnable weight matrices, and \(b_f, b_i, b_c, b_o\) represent bias terms. These parallel gate computations enable the LSTM to effectively capture long-range dependencies by selectively storing, forgetting, and outputting information over time.

\paragraph{\textbf{First LSTM Layer (Local Sequence Modeling)}}  
The first LSTM layer focuses on learning local patterns in the input sequence. 
It captures short-range dependencies between neighboring tokens. 
Suppose the input comment contains \(n\) tokens. 
This layer generates a hidden state vector for each token, written as  
\(\{h_1^{(1)}, h_2^{(1)}, \ldots, h_n^{(1)}\}\).  

Here, \(h_i^{(1)} \in \mathbb{R}^{256}\) represents the contextual embedding of the \(i\)-th token at the first LSTM layer. 
All hidden states are combined to form the matrix:

\[
H^{(1)} = [h_1^{(1)}, \ldots, h_n^{(1)}] \in \mathbb{R}^{n \times 256}.
\]

In this matrix, \(n\) is the sequence length and \(256\) is the hidden dimension size. 
A dropout rate of 0.3 is applied to reduce overfitting and improve generalization.

\paragraph{\textbf{Second LSTM Layer (Global Dependency Modeling)}}  
The second LSTM layer takes \(H^{(1)}\) as input. 
It learns long-range dependencies across the entire comment. 
This layer produces another sequence of hidden states  
\(\{h_1^{(2)}, h_2^{(2)}, \ldots, h_n^{(2)}\}\).  

Each vector \(h_i^{(2)} \in \mathbb{R}^{256}\) contains deeper contextual information than the first layer. 
The output of this layer is written as:

\[
H^{(2)} = [h_1^{(2)}, \ldots, h_n^{(2)}] \in \mathbb{R}^{n \times 256}.
\]

This hierarchical LSTM architecture enables the network to accumulate distributed contextual evidence—such as escalating hostility, sarcasm, or patterned threats—spanning across the full input sequence, thereby enhancing its ability to detect complex sequential cues.

\paragraph{\textbf{Final Sequence Representation}}  
The final sentence-level embedding is obtained from the last hidden state:
\[
h_{\text{final}} = h_n^{(2)} \in \mathbb{R}^{256},
\]
which summarizes all temporal and contextual cues essential for multilevel cyberbullying prediction.

\subsubsection{\textbf{Classification Head}}
The final sentence-level representation produced by the stacked LSTM layers is passed through a fully connected classification head to predict all cyberbullying categories simultaneously. Since a comment may express more than one form of abuse at once, the classification module is designed for multi-label outputs.

\paragraph{\textbf{Dense Projection}}  
The final hidden representation obtained from the stacked LSTM network is denoted as \(h_{\text{final}} \in \mathbb{R}^{256}\). 
Before classification, this vector is regularized using dropout with a rate of 0.3 to reduce overfitting. 
It is then passed through a fully connected dense layer to project it into a five-dimensional output space:

\[
z = W_{\text{clf}} \, \mathrm{Dropout}(h_{\text{final}},\, 0.3) + b_{\text{clf}},
\]

where \(W_{\text{clf}} \in \mathbb{R}^{5 \times 256}\) is the learnable weight matrix of the classifier, 
\(b_{\text{clf}} \in \mathbb{R}^{5}\) is the bias vector, and 
\(z \in \mathbb{R}^{5}\) represents the raw prediction logits for each abuse category. 
The output vector is written as, \[
z = [z_{\text{bully}},\, z_{\text{sexual}},\, z_{\text{religious}},\, z_{\text{threat}},\, z_{\text{spam}}].\]

\paragraph{\textbf{Sigmoid Multi-Label Activation}}  
Since the task involves multi-label classification, each category score is activated independently using the sigmoid function. 
For the \(k\)-th class, the predicted probability is computed as\cite{geeksforgeeks_sigmoid_derivative}:
\[
\hat{y}_k = \sigma(z_k) = \frac{1}{1 + e^{-z_k}}, \qquad k = 1,\ldots,5.
\]

Here, \(z_k\) is the logit corresponding to the \(k\)-th abuse label, 
\(\sigma(\cdot)\) denotes the sigmoid activation, and 
\(\hat{y}_k \in [0,1]\) represents the final predicted probability of that category being present. 
This formulation allows multiple abusive categories to be detected simultaneously, such as \text{bully + sexual}, 
\text{threat + religious insult}, or \text{bully + spam}.

\paragraph{\textbf{Binary Thresholding}}  
Final class predictions are obtained through elementwise thresholding:
\[
\tilde{y}_k =
\begin{cases}
1, & \hat{y}_k \ge 0.5, \\
0, & \hat{y}_k < 0.5,
\end{cases}
\]
producing the final multi-label cyberbullying categories.

\paragraph{\textbf{Loss Function: Binary Cross-Entropy}}  
This study uses Binary Cross-Entropy (BCE) loss because the model predicts each label separately. For multilabel detections with K labels per instance, the loss function is defined from\cite{geeksforgeeks_bce2025}: 
\[
\text{BCE} = -\frac{1}{N} \sum_{i=1}^{N} \sum_{k=1}^{K} \left[ y_{i,k} \log(p_{i,k}) + (1 - y_{i,k}) \log(1 - p_{i,k}) \right],
\]  
Here, \(N\) is the number of samples. \(K\) is the number of labels. \(y_{i,k}\) is the true label (0 or 1) for the \(k\)-th class of the \(i\)-th sample. \(p_{i,k}\) is the predicted probability for that class. The BCE loss calculates error for each label and averages it. It helps the model learn better by giving higher scores to correct labels and lower scores to wrong ones.

\paragraph{\textbf{Optimization Strategy: AdamW}}  
The full BERT–LSTM–Classifier stack is optimized using AdamW, which decouples weight decay from gradient updates. The update rule is defined as\cite{loshchilov2019adamw}:
\[
\theta_t = \theta_{t-1}
- \alpha_t \left( \frac{\hat{m}_t}{\sqrt{\hat{v}_t} + \epsilon}
+ \lambda \theta_{t-1} \right),
\]
where \(\alpha_t\) is the learning rate,  
and \(\hat{m}_t, \hat{v}_t\) are the bias-corrected moments, and \(\lambda = 0.01\) is the weight decay coefficient.

The proposed BERT--LSTM hybrid architecture is designed to perform multi-label cyberbullying classification by integrating deep contextual encoding with temporal sequence modeling. Rich contextual embeddings are first generated from Bengali text using the pre-trained BanglaBERT-Large model, which consists of 24 Transformer encoder layers and 16-head multi-head self-attention. The feature extraction procedure is outlined in Algorithm~\ref{alg:bert_internal}. The resulting contextual representations are then fed into Algorithm~\ref{alg:lstm_processing}, which implements a two-layer unidirectional LSTM network (hidden size $=256$) with inter-layer dropout ($p = 0.3$). This module captures temporal progression and long-range linguistic dependencies, producing a compact sequence representation denoted as $h_{\text{final}}$.

Finally, Algorithm~\ref{alg:bert_lstm_training} integrates all components into a unified end-to-end trainable framework. The final representation $h_{\text{final}}$ is regularized through dropout and passed to a linear classification layer that outputs five logits corresponding to the cyberbullying categories: \emph{bully}, \emph{sexual}, \emph{religious}, \emph{threat}, and \emph{spam}. The model is optimized using Binary Cross-Entropy with Logits Loss (BCEWithLogitsLoss) and the AdamW optimizer with a learning rate of $10^{-5}$, gradient clipping (maximum norm = 1.0), and a linear warmup schedule (warmup ratio = 0.1). Final predictions are obtained through a sigmoid activation, where labels are assigned using a threshold of $0.5$. Overall, the hybrid design effectively combines the deep contextual understanding provided by BERT with the sequential modeling capacity of LSTM, yielding a robust framework for multi-label cyberbullying detection in Bengali social media text.

\makeatletter
\setcounter{ALG@line}{-1} 
\makeatother
\subsection{\textbf{Performance Analysis}}
We have evaluated the performance of the models using multiple metrics, including accuracy, hamming loss, precision, recall, f1-score, mcc, cohen-kappa, roc curve, and cross-validation. Moreover,the model performance was further examined and analyzed through train-validation accuracy and loss curves, and XAI-LIME for visualization. The class imbalance was resolved by undersampling and oversampling methods in cyberbullying datasets, and the performance of balanced datasets was verified again using 5-fold cross-validation. The results of the analysis are compared thoroughly, as discussed in the next section. It is worth noting that the ”Bangla multilabel cyberbully, sexual harassment, threat, and spam detection” dataset has been used before. So, we have also conducted a comparative analysis with prior implementations available online. Therefore, we have developed a baseline framework in the field of Bangla cyberbullying detection.

\begin{table*}[t]
\centering
\renewcommand{\arraystretch}{1.2}
\resizebox{\textwidth}{!}{
\begin{tabular}{l c c c c c c c c}
\toprule
\textbf{Model} &
\textbf{Accuracy} &
\textbf{Ham.} &
\textbf{Precision} &
\textbf{Recall} &
\textbf{F1} &
\textbf{MCC} &
\textbf{Kappa} &
\textbf{AUC} \\
\midrule
KNN & 81.90 & 0.1810 & 61.41 & 69.80 & 65.34 & 37.78 & 32.78 & 75.50 \\
Naive Bayes & 87.26 & 0.1274 & 81.11 & 62.41 & 70.54 & 43.77 & 34.94 & 84.59 \\
Logistic Regression & 88.13 & 0.1187 & 81.81 & 66.10 & 73.12 & 49.43 & 44.37 & 86.67 \\
XGBoost & 88.50 & 0.1150 & 80.20 & 70.28 & 74.91 & 53.34 & 50.18 & 84.18 \\
Random Forest & 88.95 & 0.1105 & 81.46 & 70.92 & 75.83 & 54.96 & 52.56 & 86.66 \\

GRU & 91.32 & 0.0868 & 83.26 & 80.72 & 81.97 & 69.08 & 69.00 & 91.88 \\

BiLSTM & 91.34 & 0.0866 & 83.17 & 80.96 & 82.05 & 68.91 & 68.77 & 90.99 \\

BiGRU & 91.38 & 0.0862 & 82.87 & 81.61 & 82.23 & 69.60 & 69.52 & 91.60 \\

\rowcolor{lightblue}
\textcolor{skyblue}{\textbf{LSTM}} &
\textcolor{skyblue}{\textbf{91.52}} &
\textcolor{skyblue}{\textbf{0.0848}} &
\textcolor{skyblue}{\textbf{83.08}} &
\textcolor{skyblue}{\textbf{82.01}} &
\textcolor{skyblue}{\textbf{82.54}} &
\textcolor{skyblue}{\textbf{69.16}} &
\textcolor{skyblue}{\textbf{69.07}} &
\textcolor{skyblue}{\textbf{91.14}} \\
\bottomrule
\end{tabular}
}
\caption{Model Comparison on Cyberbullying Detection using Baseline Models}
\label{tab:baseline_ml_dl}
\end{table*}

\begin{table*}[t]
\centering
\renewcommand{\arraystretch}{1.15}
\resizebox{\textwidth}{!}{
\begin{tabular}{l c c c c c c c c c c}
\toprule
\textbf{Model} &
\textbf{Train Acc.} &
\textbf{Val. Acc.} &
\textbf{Test Acc.} &
\textbf{Hamming Loss} &
\textbf{Precision} &
\textbf{Recall} &
\textbf{F1} &
\textbf{MCC} &
\textbf{Kappa} &
\textbf{AUC} \\
\midrule

mBERT 
& 97.80 & 92.56 & 92.56 & 0.0744 & 85.84 & 83.29 & 84.55 & 74.30 & 74.22 & 93.36 \\

RoBERTa
& 97.08 & 92.80 & 93.46 & 0.0654 & 87.19 & 85.86 & 86.52 & 77.39 & 77.37 & 94.13 \\

BERT-BiGRU
& 98.64 & 93.48 & 93.68 & 0.0628 & 87.39 & 86.83 & 87.11 & 78.64 & 78.63 & 94.80 \\

DeBERTa
& 97.48 & 93.07 & 93.72 & 0.0632 & 85.91 & 88.67 & 87.27 & 78.18 & 78.09 & 93.95 \\

BERT-BiLSTM
& 98.40 & 93.88 & 93.84 & 0.0616 & 86.80 & 88.19 & 87.49 & 78.53 & 78.48 & 93.82 \\

BERT
& 98.82 & 93.37 & 93.92 & 0.0608 & 87.02 & 88.27 & 87.64 & 78.93 & 78.90 & 93.80 \\

BERT-GRU
& 98.77 & 93.62 & 93.97 & 0.0603 & 86.76 & 88.92 & 87.82 & 78.78 & 78.73 & 94.08 \\

indicBERT
& 98.40 & 93.46 & 94.00 & 0.0601 & 87.83 & 87.55 & 87.69 & 79.06 & 79.04 & 95.03 \\

\rowcolor{lightblue}
\textcolor{skyblue}{\textbf{BERT-LSTM}} &
\textcolor{skyblue}{\textbf{98.64}} &
\textcolor{skyblue}{\textbf{93.54}} &
\textcolor{skyblue}{\textbf{94.17}} &
\textcolor{skyblue}{\textbf{0.0583}} &
\textcolor{skyblue}{\textbf{88.35}} &
\textcolor{skyblue}{\textbf{87.71}} &
\textcolor{skyblue}{\textbf{88.03}} &
\textcolor{skyblue}{\textbf{79.40}} &
\textcolor{skyblue}{\textbf{79.33}} &
\textcolor{skyblue}{\textbf{92.66}} \\

\bottomrule
\end{tabular}
}
\caption{Model Comparison on Cyberbullying Detection using Transformer Models}
\label{tab:bert_models}
\end{table*}

\begin{table*}[t]
\centering
\renewcommand{\arraystretch}{1.15}
\resizebox{\textwidth}{!}{
\begin{tabular}{c c c c c c c c c c c}
\toprule
\textbf{Learning Rate} & \textbf{Train Acc.} & \textbf{Val. Acc.} & \textbf{Test Acc.} & \textbf{Hamming Loss} & \textbf{Precision} & \textbf{Recall} & \textbf{F1} & \textbf{MCC} & \textbf{Kappa} & \textbf{AUC} \\
\midrule
$1\times10^{-6}$ & 99.45 & 92.95 & 91.74 & 0.0826 & 79.99 & 88.27 & 83.93 & 73.14 & 72.79 & 93.93 \\
\rowcolor{lightblue}
\textcolor{skyblue}{\textbf{$1\times10^{-4}$}} &
\textcolor{skyblue}{\textbf{99.45}} &
\textcolor{skyblue}{\textbf{92.95}} &
\textcolor{skyblue}{\textbf{93.39}} &
\textcolor{skyblue}{\textbf{0.0661}} &
\textcolor{skyblue}{\textbf{87.52}} &
\textcolor{skyblue}{\textbf{85.06}} &
\textcolor{skyblue}{\textbf{86.27}} &
\textcolor{skyblue}{\textbf{76.39}} &
\textcolor{skyblue}{\textbf{76.24}} &
\textcolor{skyblue}{\textbf{93.18}} \\
\bottomrule
\end{tabular}
}
\caption{Learning Rate Comparison for the BERT-LSTM Models}
\label{tab:lr_comparison}
\end{table*}

\begin{table*}[t]
\centering
\renewcommand{\arraystretch}{1.15}
\resizebox{\textwidth}{!}{
\begin{tabular}{l c c c c c c c c c c}
\toprule
\textbf{Optimizer} & \textbf{Train Acc.} & \textbf{Val. Acc.} & \textbf{Test Acc.} & \textbf{Hamming Loss} & \textbf{Precision} & \textbf{Recall} & \textbf{F1} & \textbf{MCC} & \textbf{Kappa} & \textbf{AUC} \\
\midrule
RAdam & 99.50 & 93.48 & 93.97 & 0.0603 & 86.70 & 89.00 & 87.83 & 79.20 & 79.16 & 94.82 \\
RMSprop & 99.58 & 93.58 & 94.07 & 0.0593 & 87.04 & 89.00 & 88.01 & 79.57 & 79.51 & 95.09 \\
\rowcolor{lightblue}
\textcolor{skyblue}{\textbf{Adafactor}} &
\textcolor{skyblue}{\textbf{99.57}} &
\textcolor{skyblue}{\textbf{93.44}} &
\textcolor{skyblue}{\textbf{94.15}} &
\textcolor{skyblue}{\textbf{0.0585}} &
\textcolor{skyblue}{\textbf{86.96}} &
\textcolor{skyblue}{\textbf{89.48}} &
\textcolor{skyblue}{\textbf{88.20}} &
\textcolor{skyblue}{\textbf{79.86}} &
\textcolor{skyblue}{\textbf{79.79}} &
\textcolor{skyblue}{\textbf{94.93}} \\
\bottomrule
\end{tabular}
}
\caption{Optimizer Comparison for the BERT-LSTM Model}
\label{tab:optimizer_comparison}
\end{table*}

\begin{table*}[t]
\centering
\renewcommand{\arraystretch}{1.15}
\resizebox{\textwidth}{!}{
\begin{tabular}{l c c c c c c c c c c}
\toprule
\textbf{Technique} &
\textbf{Train Acc.} &
\textbf{Val. Acc.} &
\textbf{Test Acc.} &
\textbf{Hamming Loss} &
\textbf{Precision} &
\textbf{Recall} &
\textbf{F1} &
\textbf{MCC} &
\textbf{Kappa} &
\textbf{AUC} \\
\midrule

Original(Imbalanced)
& 98.64 & 93.54 & 94.17 & 0.0583 & 88.35 & 87.71 & 88.03 & 79.40 & 79.33 & 92.66 \\

Undersampling(Train)
& 98.02 & 90.17 & 90.42 & 0.0958 & 74.72 & 91.89 & 82.42 & 69.20 & 67.19 & 92.09 \\

\rowcolor{lightblue}
\textcolor{skyblue}{\textbf{Oversampling(Train)}} &
\textcolor{skyblue}{\textbf{99.53}} &
\textcolor{skyblue}{\textbf{93.41}} &
\textcolor{skyblue}{\textbf{94.31}} &
\textcolor{skyblue}{\textbf{0.0569}} &
\textcolor{skyblue}{\textbf{88.23}} &
\textcolor{skyblue}{\textbf{88.51}} &
\textcolor{skyblue}{\textbf{88.37}} &
\textcolor{skyblue}{\textbf{80.31}} &
\textcolor{skyblue}{\textbf{80.25}} &
\textcolor{skyblue}{\textbf{95.08}} \\

\bottomrule
\end{tabular}
}
\caption{Performance Analysis of the Proposed BERT-LSTM Model Under Different Sampling Strategies}
\label{tab:bertlstm_sampling}
\end{table*}

\begin{table*}[t]
\centering
\renewcommand{\arraystretch}{1.15}
\resizebox{\textwidth}{!}{
\begin{tabular}{c c c c c c c c c}
\toprule
\textbf{Fold} &
\textbf{Accuracy} &
\textbf{Hamming Loss} &
\textbf{Precision} &
\textbf{Recall} &
\textbf{F1} &
\textbf{MCC} &
\textbf{Kappa} &
\textbf{AUC} \\
\midrule
1 & 93.62 & 0.0638 & 83.71 & 85.31 & 84.46 & 78.13 & 77.99 & 94.91 \\
2 & 93.45 & 0.0655 & 83.41 & 83.67 & 83.41 & 77.31 & 77.20 & 93.63 \\
3 & 93.61 & 0.0639 & 85.48 & 83.06 & 84.07 & 77.65 & 77.49 & 93.63 \\
4 & 93.34 & 0.0666 & 83.81 & 83.99 & 83.84 & 77.49 & 77.45 & 93.04 \\
5 & 93.09 & 0.0691 & 83.24 & 83.37 & 83.23 & 76.63 & 76.58 & 93.79 \\
\midrule
\rowcolor{lightblue}
\textcolor{skyblue}{\textbf{Average}} &
\textcolor{skyblue}{\textbf{93.42}} &
\textcolor{skyblue}{\textbf{0.0658}} &
\textcolor{skyblue}{\textbf{83.93}} &
\textcolor{skyblue}{\textbf{83.88}} &
\textcolor{skyblue}{\textbf{83.70}} &
\textcolor{skyblue}{\textbf{77.44}} &
\textcolor{skyblue}{\textbf{77.34}} &
\textcolor{skyblue}{\textbf{93.80}} \\
\bottomrule
\end{tabular}
}
\caption{Five-Fold Cross-Validation Performance of the Proposed BERT-LSTM Model}
\label{tab:five_fold_cv}
\end{table*}

\begin{table*}[t]
\centering
\renewcommand{\arraystretch}{1.15}
\resizebox{\textwidth}{!}{
\begin{tabular}{c c c c c c c c c c}
\toprule
\textbf{Ref.} &
\textbf{Architecture} &
\textbf{Accuracy} &
\textbf{Hamming Loss} &
\textbf{Precision} &
\textbf{Recall} &
\textbf{F1} &
\textbf{MCC} &
\textbf{Kappa} &
\textbf{AUC} \\
\midrule

\cite{sunny2024deep} & Stacked LSTM
& 93.64 & -- & 94.31 & 92.05 & 91.32 & -- & -- & 98.02 \\

\rowcolor{lightblue}
\textcolor{skyblue}{\textbf{Proposed}} &
\textcolor{skyblue}{\textbf{BERT + LSTM}} 
& \textcolor{skyblue}{\textbf{94.31}} 
& \textcolor{skyblue}{\textbf{0.0569}} 
& \textcolor{skyblue}{\textbf{88.23}} 
& \textcolor{skyblue}{\textbf{88.51}} 
& \textcolor{skyblue}{\textbf{88.37}} 
& \textcolor{skyblue}{\textbf{80.31}} 
& \textcolor{skyblue}{\textbf{80.25}} 
& \textcolor{skyblue}{\textbf{95.08}} \\
\bottomrule
\end{tabular}
}
\caption{Comparison of the Proposed BERT-LSTM Model with Existing Work on the Same Dataset}
\label{tab:comparison_related_work}
\end{table*}

\begin{figure*}[htbp]
\centering
\begin{tabular}{ccc}
\includegraphics[width=0.32\textwidth, height=0.2\textheight, keepaspectratio]{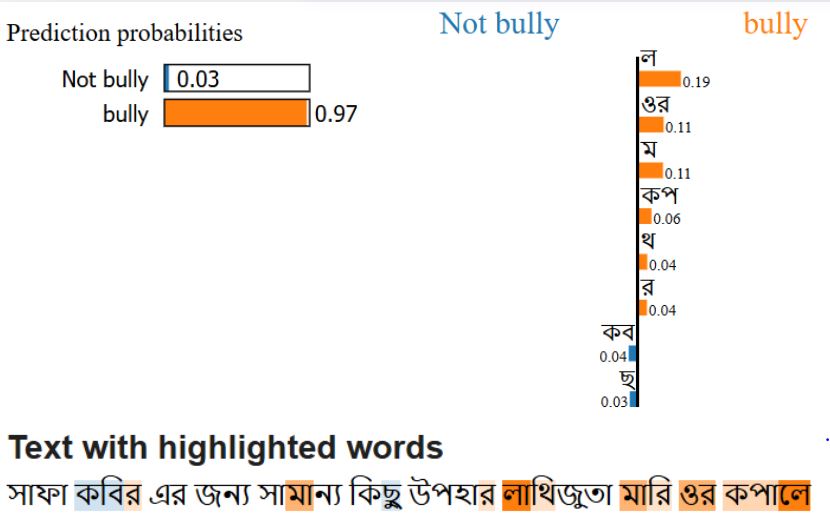} &
\includegraphics[width=0.32\textwidth, height=0.2\textheight, keepaspectratio]{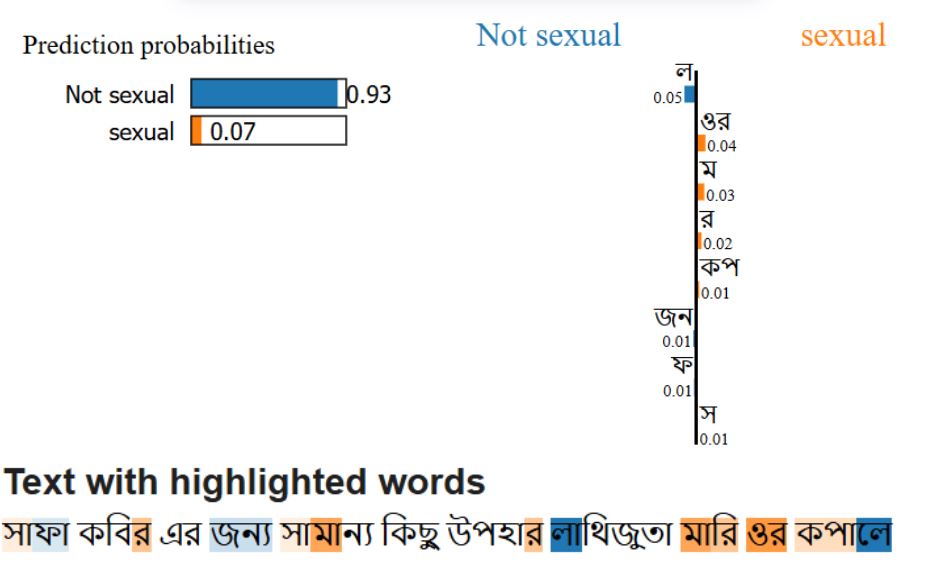} &
\includegraphics[width=0.32\textwidth, height=0.2\textheight, keepaspectratio]{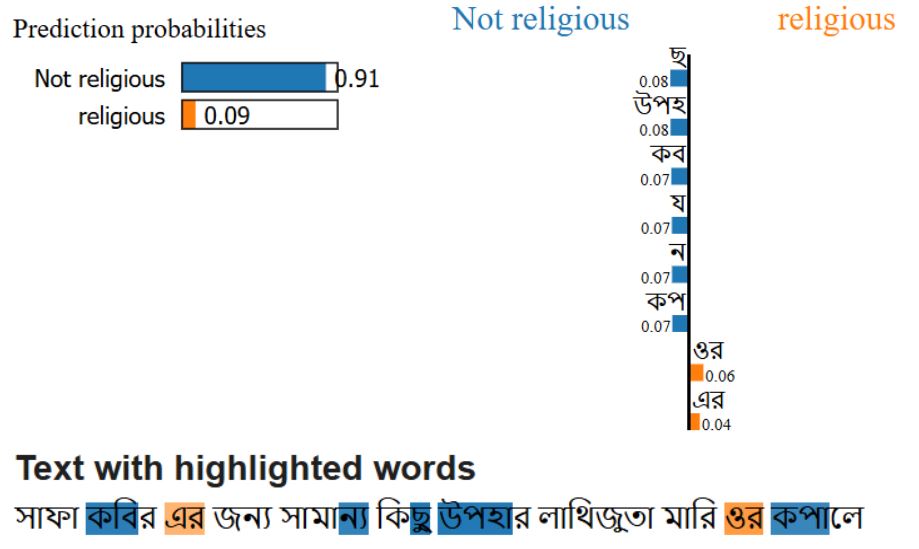} \\
(a) Bully Label Prediction &
(b) Sexual Label Prediction &
(c) Religious Label Prediction \\
\end{tabular}
\vspace{0.25cm}
\begin{tabular}{cc}
\includegraphics[width=0.32\textwidth, height=0.2\textheight, keepaspectratio]{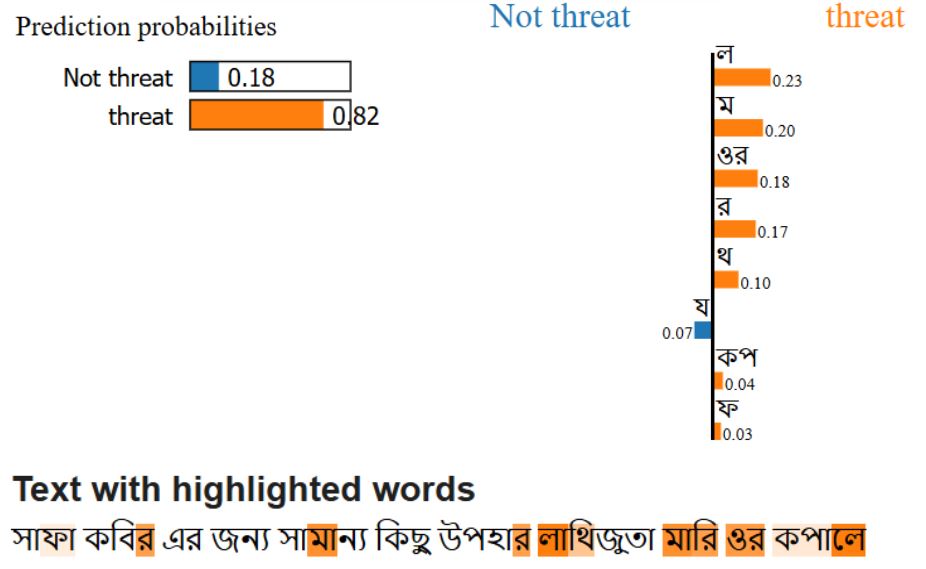} &
\includegraphics[width=0.32\textwidth, height=0.2\textheight, keepaspectratio]{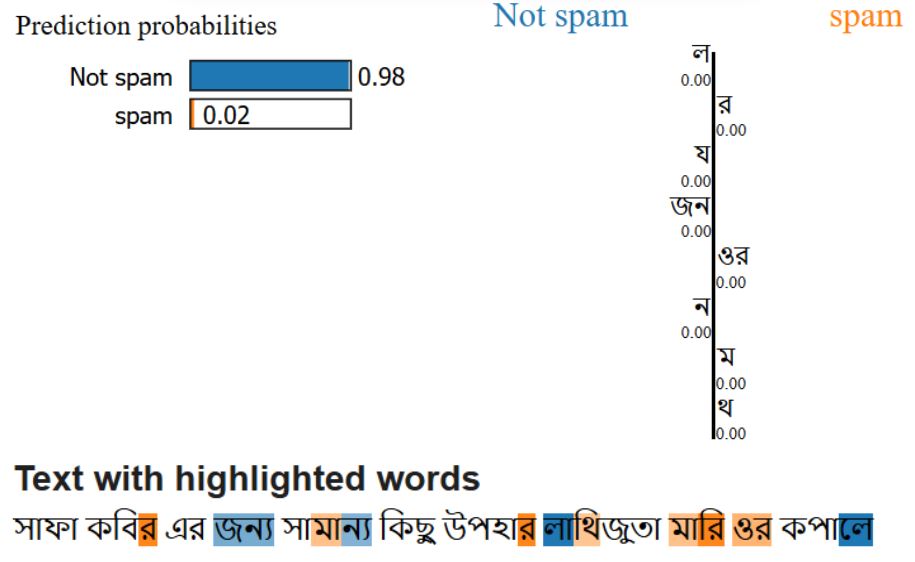} \\
(d) Threat Label Prediction &
(e) Spam Label Prediction
\end{tabular}
\caption{Multi-label Classification Predictions Showing Class Probabilities and Word-level Importance Scores for Each Cyberbullying Category. The Highlighted Words in the Bangla Text Indicate Their Contribution to the Model's Prediction for Each Label.}
\label{fig:multilabel_xai}
\end{figure*}

\section{Result Analysis}
\label{sec:4}
The experimental results of the proposed multilabel cyberbullying detection framework are presented through a series of systematic analyses. First, baseline comparisons are conducted to highlight performance gains. The impact of transformer-based contextual embeddings is then evaluated. Next, the learning rate sensitivity and the effectiveness of the optimizer are examined. The role of sampling techniques in managing class imbalance is also assessed. Finally, model robustness and stability are verified using 5-fold cross-validation on the best configuration. These analyses together offer a comprehensive understanding of the model’s accuracy, generalization, and reliability across varying experimental conditions.
Table \ref{tab:bertlstm_sampling} presents the details results of cyberbullying detection on both undersampled and oversampled. Table \ref{tab:five_fold_cv} presents the detailed results of k-fold cross-validation for cyberbullying detetction.

\subsection{Baseline Model Comparison}
Table~\ref{tab:baseline_ml_dl} presents the performance of classical machine learning and deep learning baseline models on the cyberbullying detection dataset. Classical models such as KNN, Naive Bayes, Logistic Regression, XGBoost, and Random Forest achieve accuracies ranging from 81.90\% to 88.95\%. Among them, Random Forest performs best with an accuracy of 88.95\% and an F1-score of 75.83\%. However, their relatively lower F1, MCC, and Kappa values indicate poor handling of contextual and sequential patterns in text. In contrast, deep learning models outperform classical baselines across all metrics. Recurrent architectures exceed 91\% accuracy, with LSTM achieving the highest overall scores: 91.52\% accuracy, 83.08\% precision, 82.01\% recall, 82.54\% F1-score, and top MCC 69.16\% and Kappa 69.07\% values. These results confirm that deep sequential models are more effective for cyberbullying detection by capturing temporal dependencies that classical models often overlook.

\subsection{Effect on Transformer-Based Representations}
The impact of transformer-based representations is analyzed using the results presented in Table~\ref{tab:bert_models}. The standalone BERT model demonstrates strong performance, achieving a test accuracy of 93.92\%, recall of 88.27\%, and F1-score of 87.46\%. It also records a high AUC of 93.40 and a low Hamming Loss of 0.0068, indicating strong predictive capability. However, BERT is not the top-performing model. Among the standalone transformers, indicBERT slightly outperforms BERT, achieving 94.00\% accuracy, an F1-score of 87.59\%, and the highest values for both MCC and Cohen’s Kappa at 79.04\%. These results highlight the advantage of region-specific pre-trained models for Bangla. RoBERTa and DeBERTa also deliver competitive results. RoBERTa reaches a test accuracy of 93.46\% with the highest F1-score among the baseline transformer models at 87.52\%, while DeBERTa achieves 93.64\% accuracy and 87.44\% F1. However, both models have slightly lower recall compared to BERT.

Fusion architectures like BERT-BiGRU and BERT-BiLSTM improve precision but do not show significant gains in recall or overall accuracy, with both remaining below 93.70\%. In contrast, the proposed BERT-LSTM model achieves the best overall performance. It attains the highest test accuracy of 94.17\%, F1-score of 88.03\%, and recall of 87.51\%. It also shows the lowest Hamming Loss of 0.0583, with MCC of 79.40\%, Kappa of 79.33\%, and AUC of 92.66\%. These results confirm the strength of combining BERT’s contextual embeddings with LSTM’s sequential modeling.

\subsection{Learning Rate Sensitivity}
The effect of learning rate selection on the proposed BERT-LSTM model is presented in Table~\ref{tab:lr_comparison}. A smaller learning rate of $1 \times 10^{-6}$ results in stable training but leads to lower performance, achieving an accuracy of 91.74\% and an F1-score of 83.93\%. This behavior indicates slower convergence and limited optimization progress.

Increasing the learning rate to $1 \times 10^{-4}$ results in significant improvements across all evaluation metrics. As shown in Table~\ref{tab:lr_comparison}, this configuration achieves the highest accuracy 93.39\%, precision 87.52\%, recall 85.06\%, and F1-score 86.27\%, along with improved MCC 76.39\% and Kappa 76.24\% values. These findings suggest that appropriate learning rate selection is crucial for achieving better convergence and improved generalization performance.

\subsection{Optimizer Analysis}

The impact of different optimization strategies on the proposed BERT-LSTM model is evaluated based on the results reported in Table~\ref{tab:optimizer_comparison}. Among the tested optimizers, RAdam and RMSprop achieve competitive performance, with accuracy values of 93.97\% and 94.07\%, respectively.

However, Adafactor demonstrates the best overall performance. As shown in Table~\ref{tab:optimizer_comparison}, Adafactor achieves the highest accuracy 94.15\%, precision 86.96\%, recall 89.48\%, and F1-score 88.20\%, along with the highest MCC 79.86\% and Kappa 79.79\% values. These results indicate that Adafactor provides more effective parameter updates and improved training stability for the proposed BERT-LSTM model.

\subsection{Impact of Sampling Strategies}

The effect of different sampling strategies on the proposed BanglaBERT-LSTM model is summarized in Table~\ref{tab:bertlstm_sampling}. In this study, undersampling and oversampling techniques are applied exclusively to the training data, while the validation and test sets are kept unchanged to ensure a fair and unbiased evaluation. When trained on the original imbalanced dataset, the model achieves an accuracy of 94.17\% and an F1-score of 88.03\%, representing moderate baseline performance with limited balance between precision and recall.

Undersampling slightly improves recall but does not result in a substantial overall performance gain, as many informative majority-class samples are removed. In contrast, oversampling consistently yields the best performance across all metrics. As shown in Table~\ref{tab:bertlstm_sampling}, oversampling achieves the highest accuracy 94.31\%, precision 88.23\%, recall 88.51\%, and F1-score 88.37\%, along with improved MCC 80.31\% and Kappa 80.25\% values. These findings indicate that oversampling effectively mitigates class imbalance by improving minority-class representation while preserving contextual diversity.

\subsection{Model Stability via Cross-Validation}
The robustness and generalization capability of the proposed Bangla-BERT-Large with 2-layer stacked LSTM model are evaluated using five-fold cross-validation, with the results reported in Table~\ref{tab:five_fold_cv}. The model shows consistent performance across all folds, with accuracy values ranging from 93.09\% to 93.62\%. Similar results across folds indicate that the model does not suffer from overfitting or underfitting.

As shown in Table~\ref{tab:five_fold_cv}, the averaged results achieve an overall accuracy of 93.42\%, precision of 83.93\%, recall of 83.78\%, and an F1-score of 83.70\%. The model also reports stable MCC 77.44\% and Cohen's Kappa 77.34\% values. The low variation across folds confirms that the proposed model generalizes well and maintains reliable performance under different data partitions.

\subsection{Model Interpretability Using XAI}

This study uses explainable artificial intelligence (XAI) to interpret the multilabel predictions of the proposed cyberbullying detection model. The LIME method explains each predicted label separately for the same input comment. Fig.~\ref{fig:multilabel_xai} presents visual explanations for bullying, religious, sexual, spam, and threat categories. Each subfigure highlights different words that influence the prediction of a specific label. This result shows that a single comment can activate multiple cyberbullying categories at the same time. The model assigns independent importance scores to each label. This behavior confirms proper multilabel learning. The explanations also show that the model relies on meaningful and relevant words. The XAI results increase trust and support the reliability of multilabel cyberbullying detection.

\subsection{Comparison with Current State of Work}

A comparative analysis is conducted between the proposed BERT-LSTM model and a previously reported deep learning method evaluated on the same dataset, as presented in Table~\ref{tab:comparison_related_work} to assess the effectiveness of the proposed approach. The related approach is taken from existing literature that utilizes the same experimental data and evaluation protocol~\cite{sunny2024deep}. The comparison demonstrates that the proposed framework achieves strong and well-balanced performance across multiple evaluation metrics. Specifically, the proposed model attains a classification accuracy of 94.31\% with a low Hamming loss of 0.0569\%, indicating reliable multi-label prediction with minimal misclassification.

The observed performance gains are mainly attributed to the architectural design of the proposed framework, which combines contextual transformer embeddings with sequence-aware modeling. By using Bangla-BERT-Large to capture deep semantic context and a 2-layer stacked LSTM to model word order and temporal flow, the system learns both global and local textual features essential for detecting multilabel abuse. This enables more accurate recognition of complex and overlapping abusive patterns compared to existing deep learning methods on the same dataset~\cite{sunny2024deep}. Overall, the results confirm the proposed BERT-LSTM as a strong and effective state-of-the-art solution.

\section{Conclusion}
\label{sec:5}
In this paper, we have presented a unified framework for Bangla multilabel cyberbullying detection. The proposed hybrid BanglaBERT-Large–LSTM model is evaluated on multiple dataset settings, including imbalanced, undersampled, and oversampled versions, using five-fold cross-validation and several evaluation metrics such as accuracy, Hamming loss, precision, recall, F1-score, MCC, Cohen's kappa, and ROC curves. LIME-based XAI is also applied to improve model transparency and interpretability. Experimental results show that the proposed framework consistently outperforms baseline machine learning and deep learning, as well as existing hybrid approaches. The main limitation of this work is its focus on a single language, Bangla, which limits its ability to handle code-mixed or multilingual cyberbullying content. Future work will extend this framework to support multilingual and code-mixed text multilabel cyberbullying detection, to improve real-world applicability and robustness.

\bibliographystyle{IEEEtran}

\end{document}